\documentclass[11pt]{article}

\usepackage[a4paper, margin=1in]{geometry}
\usepackage[utf8]{inputenc}
\usepackage[T1]{fontenc}
\usepackage[expansion=false]{microtype}
\usepackage{amsmath, amssymb}
\usepackage{graphicx}
\usepackage{booktabs}
\usepackage{xcolor}
\usepackage{titlesec}
\usepackage{enumitem}
\usepackage{url}
\usepackage[hidelinks]{hyperref}
\usepackage[numbers]{natbib}

\titlespacing*{\section}{0pt}{1.4em}{0.6em}
\titlespacing*{\subsection}{0pt}{1.1em}{0.4em}

\title{\textbf{What Makes a Good Terminal-Agent Benchmark Task: \\
A Guideline for Adversarial, Difficult, and Legible Evaluation Design}}

\author{Ivan Bercovich}

\date{April 2026}

\begin{document}
\maketitle

\begin{abstract}
\noindent
Terminal-agent benchmarks have become a primary signal for measuring the coding and system-administration capabilities of large language models. As the market for evaluation environments grows, so does the pressure to ship tasks quickly, often without thorough adversarial review of the verification logic. This paper is a guideline for writing good benchmark tasks, drawn from over a year of contributing to and reviewing tasks for Terminal Bench. Most people write benchmark tasks the way they write prompts. They shouldn't. A prompt is designed to help the agent succeed; a benchmark is designed to find out if it can. We argue that good tasks are adversarial, difficult, and legible, and that a large class of common failure modes --- AI-generated instructions, over-prescriptive specifications, clerical difficulty, oracle solutions that assume hidden knowledge, tests that validate the wrong things, and reward-hackable environments --- are predictable consequences of treating task authoring as prompt authoring. We catalog these failure modes, argue that real difficulty is conceptual rather than environmental, and discuss recent empirical evidence that over $15\%$ of tasks in popular terminal-agent benchmarks are reward-hackable. We hope this serves as a useful reference for benchmark maintainers, task contributors, and researchers using benchmark scores as evidence.
\end{abstract}

\section{Introduction}

Most people write benchmark tasks the way they write prompts. They shouldn't. A prompt is designed to help the agent succeed. A benchmark is designed to find out if it can.

I've been a contributor and reviewer for Terminal Bench~\citep{tbench-original} since August 2025, and this paper is about what I've learned designing and reviewing tasks. The guidance is broadly applicable to anyone building an agentic benchmark. We're currently accepting tasks for Terminal Bench 3.

I first got hooked by the challenge of making difficult tasks and seeing how long it would take for state-of-the-art (SOTA) models to catch up. I spent the most time developing the \texttt{install-windows-xp} task. The instructions the agent sees are short:

\begin{quote}
\small
Install, and run Windows XP SP3 (32-bit) in a virtual machine using QEMU. Create the virtual NTFS hard disk as \texttt{/app/isos/xp.img} to install windows. Your Windows XP ISO is at \texttt{/app/isos/xp.iso}. The administrator account should be named \texttt{tb-admin}. For your convenience, we built an API that lets you read the installation key using OCR on the CD-ROM package. You can get the official key calling \texttt{/app/read\_xp\_key\_from\_cdrom\_box}.

VNC Configuration Requirements:
\begin{itemize}[noitemsep, topsep=0pt]
    \item Configure QEMU to use VNC display \texttt{:1}
    \item Ensure VNC server is listening on port \texttt{5901}
    \item Set up a web interface (nginx) on port \texttt{80} for remote access
\end{itemize}

The VM should be left running in the background once started. You will have completed your objective when QEMU is at the Windows XP login screen and the VNC interface is accessible for monitoring.
\end{quote}

The instructions are short. But the environment was highly complex: Windows needs to be installed inside QEMU inside Docker inside Linux. The solution is tricky, since Windows XP wasn't trivially easy to install in unattended mode. The agent has to create a custom bootable ISO by extracting the original XP ISO, injecting an unattended answer file with dozens of configuration settings to suppress every possible GUI popup, adding OEM preinstallation scripts to create the required user accounts, and rebuilding the ISO with a proper El Torito boot sector. Any gaps in the unattended configuration would cause the installation to pop back into interactive GUI mode, at which point the task would have failed.

The task had a 2-hour agent timeout, one of the longest in the benchmark, because the installation alone takes 30--60 minutes under emulation. The agent has to monitor disk growth to know when the install is done. In practice this means checking every few minutes and waiting until the virtual disk exceeds 1\,GB and stops growing. Then the agent has to kill QEMU and reboot the VM in boot-only mode, without the CD-ROM attached, to reach the login screen.

It was hard for me as the task developer to be sure it was actually working, so I required the agent to set up VNC on port 5901 with an nginx web proxy so I could visually confirm the installation was progressing. For verification, I searched the virtual disk for the MD5 hashes of 12 specific Windows files: \texttt{ntoskrnl.exe}, \texttt{kernel32.dll}, \texttt{explorer.exe}, \texttt{ntldr}, and others. But I went further. The test takes a VNC screenshot and compares it against reference screenshots of the login screen using structural similarity (SSIM), requiring at least 85\% match. It also verifies the \texttt{tb-admin} user was created by searching the virtual disk for the user account picture bitmap, which only exists if the OEM setup scripts ran successfully during installation.

This task didn't end up in the official Terminal Bench 1.0 dataset because it took too long to run and made logistics painful. I think it's still one of the coolest TBench tasks created to date. I went from knowing very little about benchmarks to being an official reviewer, and submitted many other tasks which did get merged, including \texttt{install-windows-3.11} and \texttt{video-processing}.

What follows are some personal opinions about what constitutes a good task. Official guidelines are also available in the Terminal Bench 3 repository.

\section{Benchmarks Should Be Adversarial, Difficult, and Legible}

When you prompt an LLM, you want it to succeed. You repeat yourself, you emphasize, you add examples, you structure everything just right. That's what works.

A benchmark is adversarial. We tend to write instructions in a way that we are trying to encourage the agent to get it right, but that's not the point of the benchmark. The point of a benchmark with verifiable rewards is to state an unambiguous objective, which can be confidently verified, and which corresponds to a difficult task. Think of it as an interview question for a principal engineering role. Hints are for entry-level employees where you want to know if they can think well. Principal engineers need to deliver the answer on their own.

The ideal task can be described in two paragraphs. It's difficult to solve. The agent will have to think before even doing anything. But then the actual solution doesn't necessarily have to be itself very complex (e.g.\ a long program). It's not hard because it requires a lot of resources, or because the task is expansive. You can ask an agent to optimize a small program and that might be just as hard as optimizing a larger one if the requirements are aggressive enough. While not a requirement, the most elegant tasks have short, well-specified, self-explanatory (e.g.\ no README needed) instructions. Think literate programming.

A benchmark has to straddle a balance between being as realistic as possible while still remaining legible. As capabilities and time horizons expand, this becomes difficult. Running a benchmark of tasks involving swarms replicating highly complex software remains expensive, and relies on the credibility of the reviewer (in the case of Anthropic's C compiler effort, Nicholas Carlini~\citep{carlini-ccompiler}). For the rest of us, if we want our benchmarks to be credible and gain traction, we benefit from making our tasks tractable. Likewise, if we want a busy leaderboard, we have to make the benchmark accessible to all model and agent developers, which means the cost and infrastructure complexity can't be disproportionate.

\section{A Taxonomy of Bad Tasks}

To a large degree, the value gained by contributing to a project like Terminal Bench is access to concrete feedback. In that light, I want to share some specific examples of common issues.

\subsection{AI-Generated Instructions}

The most common and most obvious problem. Someone asks an LLM to write their task instructions and submits whatever comes out. It's immediately recognizable: the tone is wrong, it's verbose, it's over-structured, and it reads like it was written to maximize the probability that the agent succeeds.

Great instructions are written by hand, or heavily edited from whatever an LLM suggests. They are not meant to be a forced prompt with emphasis and repetition to coerce the attention heads to listen to you. Direct and to the point. Specific and sufficient, but not redundant and attention-grabbing.

\subsection{Over-Prescriptive Instructions}

Even when instructions are human-written, they're often too prescriptive. Authors tell the agent how to solve the problem instead of what the end state should be.

I have an aversion to this sort of instruction. It's too clerical, asking for a very specific set of steps, instead of articulating a goal. Couldn't you just describe how the system should work and let the agent figure it all out? You don't need to explain how things might fail. This is not some educational problem that a college student needs to learn from. Assuming an experienced engineer will understand what you mean, you can expect the agent to bring the same understanding.

Write the instruction as if it's intended for a smart human. Clear, but not redundant. I don't think you need to tell the agent how it will be tested. Just tell it what you expect the end result to be. But the instructions should be sufficiently specified so that meeting them implies passing the tests. Unlike when designing a prompt for high probability of success, here it's enough to say things once, as long as it's clear.

My opinion here is likely stronger than the median TBench reviewer. The reason is that I see every token as an opportunity to mistakenly add ambiguity or create a specification detail which the tests might miss. If I want the task to be unambiguous and perfectly verified, then brevity is an important KPI.

\subsection{Clerical Difficulty}

I've been thinking about the distinction between tasks that fail because the structure of the expected output is convoluted --- a long instruction explaining how to form a complex JSON object --- versus tasks that fail for something intrinsically difficult. There's something categorically different between the two. I call these clerical or administrative errors, and tasks that fail exclusively because of that usually are hard in an uninteresting way.

If an agent fails because it put a dollar sign in the amount field when you expected a float, or because it used a top-level key when you wanted a bare array, it's measuring format compliance. Using so much of the instruction to detail the output format tends to make models fail not because of the complexity of the task, but because of some clerical error. Of course it's important for models to get formatting right, every time, even when the distinction is nuanced. It's just that this is not the same sort of capability that Terminal Bench is setting out to solve. If a task is challenging to SOTA models, it shouldn't be because they can't spell ``strawberry''.

A related problem is tasks that are too wide, asking for a lot of little deliverables instead of solving a concrete large problem.

\subsection{Solutions That Assume Hidden Knowledge}

The submitted \texttt{solution.sh} should solve the problem as an agent would. Hardcoding an answer which implies knowledge not self-evident in the instructions is not helpful. The person creating the task is making assumptions that are not included in the instructions. The task is underspecified, and you don't realize until you actually read the solution.

I watch for this carefully. I want to make sure the solution doesn't display inside knowledge that the agent wouldn't be able to know. Are there some commands you can include that would reveal the exact issue before you apply the patch? A proper oracle will actually ask questions of the system. It will go through a series of commands to figure out what's wrong, and then upon figuring out what's wrong, it will produce a solution. If the solution jumps straight to the answer without exploratory steps, it might be making unfair assumptions.

The problem is that the author knows the ground truth. So it might seem that you are being fair because your invoice processing solution is looking for Levenshtein distance and so on, but you could have just as easily come up with typos that didn't pass your particular choice of data cleaning. The solution isn't investigating the issue. It's just writing down the answer to a known issue.

\subsection{Tests That Validate the Wrong Things}

Tests should verify outcomes, not implementations.

Why test that Pandas is installed? This was never requested in the task. Doing a lot of string comparisons to evaluate source code is going to be brittle. You should test examples beyond the one you tell the agent to test with.

I've seen this often: tests that check for specific libraries instead of whether the output is correct, tests so tightly coupled to the oracle solution that any alternative correct approach would fail. There was a task which required setting Linux permissions in such a way that only certain people could perform certain operations. There was one set of tests verifying functionally that the right people can do the right operations. There was another set of tests that looked at Linux permissions directly. To what degree are these testing the same thing? Is it possible for permissions to look slightly different and still accomplish the goal?

For tasks with inherent variance, testing gets harder. I ran my own \texttt{word2vec} task 25 times. The \emph{queen $-$ woman $+$ man $=$ king} analogy returned \emph{king} among the top 75 results 100\% of the time, but the variance was wide, from 1 to 40. I could test a number of things about the resulting model, but I never felt satisfied with this canonical test.

Sometimes, the task includes dependencies that the agent has access to. For example, when the goal is to implement a missing feature in a larger application. In those cases, it's important to test the outcome with a copy of the original code, to guarantee the task is not passing due to spurious changes.

What about LLM-as-a-judge? The TB3 rubric allows it in rare and extraordinary circumstances, but only if it can be shown that the verifier essentially never makes a mistake, for example, by showing that multiple different LLMs always make the same determination.

\subsection{Reward Hacking and Environment Leakage}

Make sure the agent doesn't have access to data that shouldn't be available. Anything generated or copied in the Dockerfile will generally be accessible. If you place something in \texttt{/tests}, it will be copied later, right before tests run. If the agent does have access to the reference answer, what stops it from copying it?

Because it's easier to reward hack as root, there have been discussions about root vs.\ userland agents, and TBench supports both. But limiting the agent permissions is one way to make a hard task in uninteresting ways. In my experience, it's better to give the agent unrestricted access and avoid wasting time navigating permissions.

It's essential to make sure the environment is not reward hackable. It doesn't matter if the models you're testing play by the book. Benchmarks have to be resistant to being gamed, or they are unworthy. When a popular benchmark is discovered to be hackable, the authors lose credibility, alongside many papers which might have used that benchmark as evidence of results~\citep{krakovna2020specification, metr2025hacking}.

One idea I think is helpful: modify the harness to include test signatures in the prompt and see if agents are more successful than expected. Run a ``please hack'' version and analyze those trajectories for hack success. This tells you which tasks are vulnerable before they go live. I recently implemented an automated version of this for all TBench PRs, which informed the Terminal Wrench dataset~\citep{terminal-wrench-2026}: a systematic audit of five public terminal-agent benchmarks found that over 15\% of tasks were demonstrably reward-hackable, with exploits ranging from simple output spoofing to standard-library patching and rootkit-style binary hijacking.

\section{What Difficulty Actually Means}

In general, difficulty should come from the problem, not from the environment. Not from resource pressure, not from verbose instructions, not from trick formatting. There is a true measure of difficulty, and we are sort of trying to come up with a way to estimate it. We may be off. Even our own judgment may be off.

How do you estimate the difficulty? Did you test it against various models? What were the results? How many SOTA agents did you try? Can you show a few failure examples and \texttt{harbor task debug} output for them? If you can't show interesting failures from capable models, your task might be short of great.

What counts as ``hard'' is an ongoing discussion. The TB3 rubric anchors difficulty on what's hard for a human expert, and that's a reasonable starting point. But LLM capabilities are very jagged~\citep{jagged-frontier}. The analogy of something being harder for humans as also being a challenge to LLMs does not always hold. LLMs can read and cross-reference a million lines of logs in seconds. That's not hard for them even though it would take a person days. Conversely, things humans find straightforward, like navigating an interactive TUI, can completely stall an agent. So testing against models isn't the definition of difficulty, but it's the best diagnostic we have, and it often reveals that what you thought was hard is actually easy, or vice versa.

The difficulty bar keeps rising as SOTA improves. Many tasks I reviewed for TB2 that seemed reasonable would be too easy today. The TB3 rubric is explicit:

\begin{quote}
As a general rule, any task that could be solved by an average undergraduate student in under a few days is too easy. This means typical problems that feel like course projects are unlikely to be accepted. These projects include tasks like writing simple compilers or operating systems; or implementing simple algorithms or protocols.
\end{quote}

There might be a natural timeout beyond which more reasoning is hopeless, but artificially low timeouts hide insights. I suspect a lot of useful information is going into the timeout black hole. It would be useful for tests to be progressive --- giving a letter grade to the agent solution --- to see a continuous tradeoff between time spent and quality.

We tried letting the agent know about its timeout: ``you have 10 minutes left, you have five minutes left.'' I thought that would work, but actually it makes the agents freak out towards the end, and they start doing irrational stuff. So that didn't work.

Constraining resources creates a different kind of difficulty, and often an unfair one. If agents assume normal conditions and we constrain resources without putting it in the prompt, that's unfair. It's also hard to get exactly right. Tasks that pass in one infrastructure setting will fail in an almost identical counterpart when resources are intentionally a limiting factor.

Making a task harder by making it bigger --- extracting more images, adding more of the same but longer --- does not make it better. Real difficulty is conceptual. Can the agent figure out the approach? Can it debug when things go wrong? Can it reason about the problem before diving into execution?

A task that takes 30 minutes because the agent is wrestling with the problem is more interesting than one that takes 30 minutes because \texttt{make -j4} is running.

\section{Building Tasks That Work}

\textbf{Run it yourself.} When I get stuck debugging a task, I run it with the oracle, interact with the container, try to run the tests with \texttt{docker exec} and see what happens. If the failure is from an agent solution, convert the agent log into an alternative \texttt{solution.sh} and step through it. In our runtime for testing, long-running processes will sometimes cause the harness to wait until a timeout before running the tests. During development, it's useful to be inside the container and avoid wasting time.

\textbf{Watch agents fail.} Something that is very helpful is looking at logs of agents trying to solve the task, particularly those that fail, and trying to understand why. When it fails, you have to figure out: did they fail because it's hard, or did they fail because it's unfair? Did it fail because the instructions were insufficient? Because the tests were overly aggressive? Or did they actually fail because they didn't know what to do?

We need to convince ourselves that failures are not because of a deficiency in the task. Run in batches of 5, then use the debug command. You might be skeptical that GPT-5.4 is failing at some task. It seems easy and one-shot. Can you test it with 5 trials and then see what the failures have in common? There was an instance where GPT-5 liked to use \texttt{nano} for editing files, but then it opens it and can't use it. It doesn't know what to do. It gets stuck in interactive mode and fails. These are the kinds of insights you get from watching trajectories.

\textbf{Solutions should be deterministic, but the problem can still be dynamic.} One area where people tend to fork is: should the AI give you the answer, or should it give you a piece of software that gives you the answer? In my experience, the latter tends to be more verifiable. It's much more testable. Tasks should be graded on outcomes, not process. As the TB3 rubric puts it: ``a task cannot say `use emacs to edit a file' --- using vim is allowed. The only exception is verification that prevents cheating.''

\section{Why This Matters}

AI is a very empirical field. Without hands-on experience it's hard to develop the right intuition. When we started soliciting tasks for TB3, we rejected the first several proposals. Then, like the four-minute mile, once a good PR came through, many more followed.

Benchmarks are where SOTA has to earn its name. This is where you develop your intuition for what we might see in the coming months and years. Anyone can get access to AI. Most people don't have a good sense for where we are in the capability curve.

Tasks should be authentic engineering problems, not artificial constructs. The best ones come from real problems someone actually had to solve. It's kind of crazy that major labs are outsourcing semi-artificial tasks when people are doing real tasks every day. People are doing things with LLMs that could be tasks without realizing so. Every week you spend a few hours solving a problem, that's a task.

The best tasks describe a real problem that an experienced engineer would recognize, in language that an experienced engineer would use, with tests that verify the outcome rather than the approach. I suspect we'll start seeing difficult tasks emerge from the vibecoding world, particularly for complex functions like finance, where the problem is genuinely hard but the person who needs it solved isn't a developer. The scope of what we should test is wider than we think.

\section*{Acknowledgments}

Thanks to the Terminal Bench maintainers and contributor community, and to the reviewers who provided feedback on tasks discussed here.

\end{document}